# COMPARATIVE STUDY OF SENTIMENT ANALYSIS FOR MULTI-SOURCED SOCIAL MEDIA PLATFORMS

Keshav Kapur[1] and Rajitha Harikrishnan[2]

[1]Department of Information and Communication Technology, Manipal Institute of Technology, Manipal, India
keshav29kapur@gmail.com

[2]Department of Information and Communication Technology, Manipal Institute of Technology, Manipal, India
rajithasuja@gmail.com

## ABSTRACT

*There is a vast amount of data generated every second due to the rapidly growing technology in the current world. This area of research attempts to determine the feelings or opinions of people on social media posts. The dataset we used was a multi-source dataset from the comment section of various social networking sites like Twitter, Reddit, etc. Natural Language Processing Techniques were employed to perform sentiment analysis on the obtained dataset. In this paper, we provide a comparative analysis using techniques of lexicon-based, machine learning and deep learning approaches. The Machine Learning algorithm used in this work is Naive Bayes, the Lexicon-based approach used in this work is TextBlob, and the deep-learning algorithm used in this work is LSTM.*

## 1. INTRODUCTION

The rise of the internet has altered how people now express their ideas and thoughts. These days, people tend to express their opinions largely on social networking websites like Twitter, Reddit, etc. Online communities provide us with interactive media where users can use forums to inform and persuade others. Additionally, social media gives businesses a chance by offering them a platform to engage with their customers for advertising. The inference of user sentiment can be quite helpful in the field of recommender systems and personalization to make up for the absence of clear user input on a given service.

Automatic sentiment analysis (SA) is a problem that is becoming popular in the research domain. Sentiment analysis can be defined as a process that automates mining of attitudes, opinions, views and emotions from text, speech, tweets and database sources through Natural Language Processing (NLP)[1]. Sentiment analysis involves classifying opinions in text into categories like "positive" or "negative" or "neutral". Despite the importance of SA and the variety of applications it has now, there are a number of difficulties with natural language processing that must be overcome. SA has been used in various domains such as commercial products, feedback systems, social media, etc. In the past few years, sentiment analysis for social media platforms has been platform specific. In this paper our objective was to understand the divergence and optimality of models trained on datasets sourced from multiple social media platforms. As we are aware that language styles, age demographics, geographical locations vary drastically in all social media platforms, we chose two platforms with unique identities, i.e, Reddit and Twitter. The user base of these platforms have been found to be non coinciding and due to the nature of communication in these platforms, we were able to continue with our approach with non-overlapping data subsets. We have selected three algorithms to select the best classification technique common for all the social media platforms: Naive Bayes Classifier, TextBlob[5] with Random Forest Classifier and a Bidirectional Long Short-Term Memory (BiLSTM) based classifier[2].

## 2. PREPROCESSING

All the Tweets and Comments (from Twitter and Reddit Comment Section) were extracted using Twitter and Reddit APIs, i.e, Tweepy and PRAW respectively. The Tweets and Comments were pre-processed using Python re (Regex Library) with a Sentimental Label to each ranging from 1, 0 and -1 as shown in Table 1.

| Sentiment | Label |
|---|---|
| Positive Tweet/Comment | 1 |
| Neutral Tweet/Comment | 0 |
| Negative Tweet/Comment | -1 |

Table 1. Sentiment Labels

The dataset has 200118 data points divided into 3 categories (1, 0, -1). The categorical distribution of sentimental labels in the dataset is shown in Figure 1.

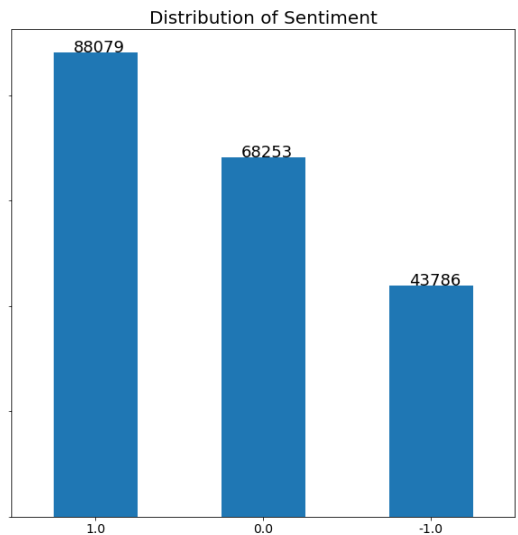

Figure 1. Distribution of Sentiment

For preprocessing, we removed URLs, emoticons, hashtags, and mentions using a python package *tweetpreprocessor*. After that we removed: contractions from the tweets, special characters and extra spaces. Then we used a python package called Natural Language Toolkit (NLTK) for removing the stop words. [3]

## 3. MODELS

In this work, we wanted to highlight the differences of using normal NLP techniques, machine learning algorithms, and deep learning techniques. Hence, we have chosen TextBlob: a rule-based approach [4] which uses simple NLP technique, Naive Bayes: a probabilistic

Machine Learning algorithm, and LSTM: a deep learning algorithm which makes use of RNNs. We have mainly used three techniques for sentiment analysis on our dataset:

### 3.1. TextBlob: a Lexicon-based method

Textblob is a python library for Natural Language Processing which is actively used to achieve certain tasks. A sentiment is identified by its semantic orientation and the intensity of each word in the sentence for lexicon-based techniques. This approach requires a dictionary which has a collection of positive and negative words. A text message is represented by a bag of words. Individual scores are assigned to each word after which final sentiment is calculated by performing the average of all scores.

TextBlob returns two parameters namely polarity and subjectivity. Polarity measures how positive or negative the given problem instance is. Its value lies between [-1, 1], -1 indicates a negative statement and 1 indicates a positive statement. Subjectivity quantifies the opinions or perspectives that need to be examined in the context of the problem statement. Its value lies between [0, 1], 1 indicates high subjectivity which means that the text contains personal opinion rather than factual information and 0 indicates low subjectivity. Subjectivity is calculated with the help of another parameter present in TextBlob called *intensity* which implies if a word modifies the next word.

After obtaining the polarity and subjectivity, a Random Forest Classifier model was used to make the predictions. It is a machine learning algorithm. Decision Trees are the building blocks of Random Forest Classifier. In our model, the predictions are based on 25 Decision Trees which were passed as a hyperparameter for Random Forest Classifier. Each individual tree in the random forest gives a prediction class. The class with the most votes becomes our model's prediction.

### 3.2. Naive Bayes Classifier

Naive Bayes Classifier is a Machine Learning Algorithm which makes use of probability to make predictions based on prior knowledge of conditions. Assumptions made in Naive Bayes is that each feature can make an equal and independent contribution to the outcome. Preprocessing is performed on the 'text' column of our dataset as there is a need for our text data to convert into numeric type for probability calculations which helps in prediction. The probability is calculated on the basis of Bayes Theorem and Conditional Probability. Bayes Theorem is given by:

$$P(A|B) = P(B|A) \times P(A)/P(B)$$

where A and B are two independent events. After pre-processing the dataset, a document-term matrix is constructed. Every distinct word in the corpus's lexicon produces a new feature, which causes the document-term matrix to produce a large feature space. To reduce the dimensions and enhance the model performance, we perform the data cleaning steps. TF-IDF method is used to represent a particular text in the form of a vector. N-grams are used to understand the context of the text.

### 3.3. Bidirectional Long Short-Term Memory Model (BiLSTM)

A Long Short-Term Memory model, or LSTM, is a type of Recurrent Neural Network (RNN) used to process temporal data. Each line has features that furnish context when read in order, which is why we have chosen a Bidirectional LSTM model. Before training the model, we tokenise our input sentences into sequences of integers and then pad each sequence to the same length. For tokenization we use the *Keras* Tokeniser class which creates a vocabulary index based on word frequency and replaces each word corresponding to the vocabulary index integral value. After that, we pad the final sequence of integers with '0'. Then we split the dataset into Training, Validation and Test Set.

In this work, BiLSTM is used to enable additional training by crossing over the input data sequences twice. This can be done by including two LSTM layers, one takes the forward direction sequences, and the other takes the backwards direction sequences, which offers better predictions. In this way, The approach can capture contextualised word representations in a clause. The goal of BiLSTM is to model (1) complex semantic and syntactic characteristics of words (2) lexical ambiguity or polysemy, words with similar pronunciations could have different meanings at different locations or contexts. Our proposed model has the architecture as shown in Figure 2. We tuned the model with the following hyperparameters: epochs=20, learning rate=0.1, momentum=0.8. We trained the model with the batch size of 64 and chose categorical cross entropy loss function with Stochastic gradient descent optimizer.

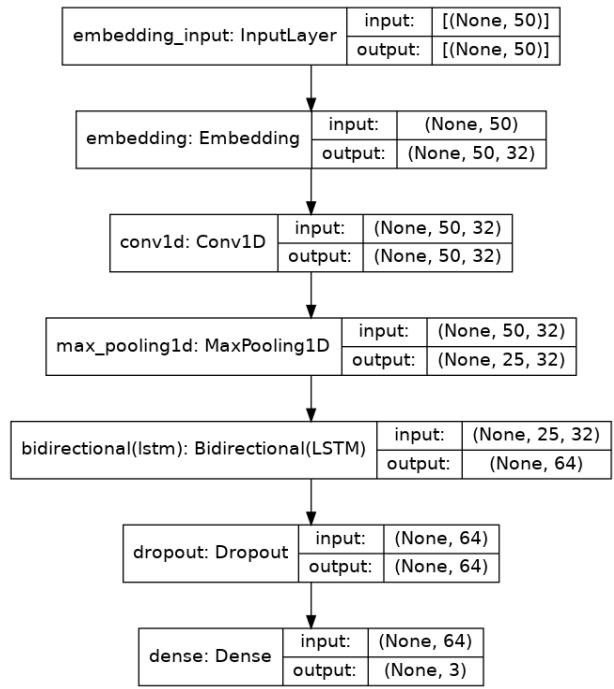

Figure 2. LSTM Model Architecture

## 3. EVALUATION

In the testing phase, we tested over 25% of our dataset which is approximately 50033 data points. The results of the test have been published in Table 2. Table 2 shows Accuracy, Precision, Recall and F1-Score of each model. As it can be seen from the table, the custom BiLSTM had the best performance followed by the TextBlob Lexicon Based model and Naive Bayes Classifier.

| Model | Accuracy | Precision | Recall | F1 Score |
|---|---|---|---|---|
| Naive Bayes | 0.6968 | 0.7220 | 0.6968 | 0.6900 |
| TextBlob + Random Forest | 0.8159 | 0.8193 | 0.8152 | 0.8106 |
| Custom BiLSTM | 0.8963 | 0.8986 | 0.8912 | 0.8949 |

Table 2. Evaluation Scores

## 3. CONCLUSIONS

In this paper, we show a comparative study of existing techniques for sentiment analysis using lexicon-based approach, machine learning approach and deep learning approach. In the lexicon-based approach, we have used Random Forest Classifier along with it as the dataset we considered was imbalanced and decision trees perform better on such datasets. In the deep learning approach, we have used BiLSTM as each component in the input sequence has information about both the past and present which helps in providing us with more optimal output. It was observed from the results that TextBlob along with Random Forest Classifier and Custom BiLSTM have performed well. Our future work would involve understanding the data characteristics which affect the performance of Naive Bayes Classifier, training our model in RoBerta

## Authors

Keshav Kapur

Currently pursuing his B.Tech in Computer and Communication Engineering at Manipal Institute of Technology. His key interests lie in Natural Language Processing, Computer Vision and Robotics. He is an open source enthusiast and is currently working on real world applications of Machine Learning.

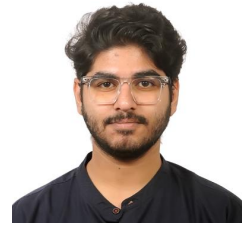

Rajitha Harikrishnan

Currently pursuing her B.Tech in Computer and Communication Engineering at Manipal Institute of Technology. Her key interests lie in Natural Language Processing, Data Science, and Data Analytics.

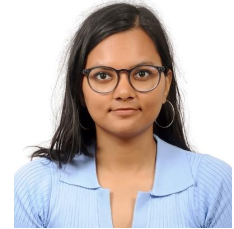